\documentclass[journal]{IEEEtran}
\ifCLASSINFOpdf
\else
   \usepackage[dvips]{graphicx}
\fi
\usepackage{url}
\usepackage{graphicx}
\usepackage{amsmath}
\usepackage{amssymb}
\usepackage{graphicx}
\usepackage{caption}
\usepackage{xcolor}
\definecolor{darkgreen}{rgb}{0.0, 0.5, 0.0}
\usepackage{subcaption} 
\usepackage{cite}
\usepackage{hyperref}
\usepackage{float}
\begin{document}

\title{Pathology-Aware Multi-View Contrastive Learning for Patient-Independent ECG Reconstruction}

\author{Youssef Youssef and Jitin Singla
\thanks{Y. Youssef is with the Department of Computer Science and Engineering, Indian Institute of Technology Roorkee, Roorkee 247667, India (e-mail: youssef\_yy@cs.iitr.ac.in).}%
\thanks{J. Singla is with the Department of Biosciences and Bioengineering, Indian Institute of Technology Roorkee, Roorkee 247667, India (e-mail: jitin.singla@bt.iitr.ac.in).}%
\thanks{*Corresponding author: Jitin Singla.}}

\markboth{}
{Shell \MakeLowercase{\textit{et al.}}: Bare Demo of IEEEtran.cls for IEEE Journals}
\maketitle

\begin{abstract}
Reconstructing a 12-lead electrocardiogram (ECG) from a reduced lead set is an ill-posed inverse problem due to anatomical variability. Standard deep learning methods often ignore underlying cardiac pathology losing vital morphology in precordial leads. We propose Pathology-Aware Multi-View Contrastive Learning, a framework that regularizes the latent space through a pathological manifold. Our architecture integrates high-fidelity time-domain waveforms with pathology-aware embeddings learned via supervised contrastive alignment. By maximizing mutual information between latent representations and clinical labels, the framework learns to filter anatomical "nuisance" variables. On the PTB-XL dataset, our method achieves approx. 76\% reduction in RMSE compared to state-of-the-art model in patient-independent setting. Cross-dataset evaluation on the PTB Diagnostic Database confirms superior generalization, bridging the gap between hardware portability and diagnostic-grade reconstruction.
\end{abstract}


\IEEEpeerreviewmaketitle
\section{Introduction}
\IEEEPARstart{T}{he} standard 12-lead electrocardiogram (ECG) is the clinical gold standard for assessing cardiac electrical activity \cite{mason2024,Sattar2023ECG,SCST2024ECG}, yet its utility is constrained by the requirement for ten electrodes and precise anatomical placement, making it challenging for continuous monitoring in ambulatory and home settings \cite{SCST2024ECG,moghaddam2025,Giannetta2020ECGPlacement,Batchvarov2007ECG}. To bridge the gap between clinical interpretability and hardware portability, there is a growing need to reconstruct high-fidelity 12-lead waveforms, especially precordial $V1-V6$ leads, from a reduced set of measured leads (e.g., Leads I, II, and $V_2$) \cite{moghaddam2025,Obianom2025ECGReconstruction}. The feasibility of ECG lead reconstruction is theoretically grounded in the spatial redundancy of cardiac dipoles \cite{moghaddam2025,Jain2023ECGRedundancy,nelwan2005}. However, the relationship between the cardiac source $\mathbf{v}(t) \in \mathbb{R}^3$ and body surface potentials $\mathbf{\Phi}$ is governed by a non-linear subject-specific lead-field matrix $\mathbf{A}(\mathbf{s})$, where $\mathbf{s}$ represents latent anatomical variables such as torso geometry, heart orientation and tissue conductivity \cite{Buist2002,Ushenin2021}:
\begin{equation}
\mathbf{\Phi}(t) = \mathbf{A}(\mathbf{s})\mathbf{v}(t) + \epsilon
\end{equation}

In a patient-independent setting, $\mathbf{s}$ is unobserved. Consequently, standard models attempting a direct mapping $f: \mathcal{X} \to \mathcal{Y}$ are mathematically forced to minimize error by learning the marginalized transfer function $\mathbb{E}[\mathbf{A}(\mathbf{s})]$. This is expressed as:
\begin{equation}
p(\mathcal{Y} | \mathcal{X}) = \int p(\mathcal{Y} | \mathcal{X}, \mathbf{s}) p(\mathbf{s} | \mathcal{X}) d\mathbf{s}
\end{equation}

Because the influence of $\mathbf{s}$ is integrated (averaged) across the population, the model suffers from a "regression to the mean" effect, where high-frequency diagnostic details are lost to satisfy the population average. This loss is particularly high in the precordial "transitional zone" ($V_3$–$V_4$) where anatomical variance is maximal \cite{Presacan2025ECGReconstruction}.

To counteract this marginalization, we propose Pathology-Aware Multi-View Contrastive Learning, a framework that reformulates lead reconstruction by regularizing the latent space through a pathological manifold. We argue that while the anatomical matrix $\mathbf{A}(\mathbf{s})$ varies stochastically between patients, the cardiac signal can be constrained to lower-dimensional manifold dictated by the underlying pathology $l$. Since $\mathbf{s}$ is inaccessible for new patients, we introduce a pathology-aware representation $\mathbf{h} = f_\phi(\mathcal{X})$ to act as a structural anchor and conditioning variable, decomposing the generative process as:\begin{equation}p(\mathcal{Y} | \mathcal{X}, \mathbf{h}) \propto p(\mathcal{Y} | \mathcal{X}, l) p(\mathbf{h} | \mathcal{X}, l)\end{equation}

We maximize the Mutual Information $I(\mathbf{h}; l) = H(l) - H(l|\mathbf{h})$ between latent representation $\mathbf{h}$ and pathology label $l$ by optimizing a supervised contrastive loss. This partitions latent space by clinical condition rather than anatomical noise, providing a prior that restricts the solution space for $\mathcal{Y}$ to the "pathological subspace". Mathematically, this conditioning reduces the conditional entropy of the reconstruction as $H(\mathcal{Y} | \mathcal{X}, \mathbf{h}) < H(\mathcal{Y} | \mathcal{X})$, providing the reconstruction decoder with a semantic "anchor" to recover features typically lost in unconditioned models.

Our contributions are two-fold: 1) We propose a patient-independent multi-view architecture that fuses high-fidelity waveforms with contrastive embeddings; 2) We achieve state-of-the-art results on PTB-XL Dataset, including 76\% reduction in RMSE over existing benchmarks, and demonstrate robust cross-dataset generalization on the PTB Diagnostic Database.

\section{Related Work}
Early ECG reconstruction methods used linear transformation techniques \cite{finlay2007} (e.g., the inverse Dower matrix \cite{Edenbrandt1988InverseDower}, regression-based models \cite{Kors1990FrankVCG}) to model spatial relations between leads. These mathematical models were interpretable; however, their precision is highly affected by variations in human anatomy and electrode placement \cite{Obianom2025ECGReconstruction,nelwan2005}. To overcome these limitations, recent research has changed direction to using deep learning (DL) methods to model the non-linear relationship between leads from the data directly. There are several common architectures reported in the literature, including:
\begin{itemize}
     \item U-Net architectures: These have been used to reconstruct a full 12-lead ECG from a single input lead, achieving correlations up to 0.9 for specific leads \cite{Beco2022LeadConversion}.
    \item Hybrid CNN-LSTM Models: One-dimensional Convolutional Neural Networks (CNN) combined with Bidirectional Long Short-Term Memory (LSTM) models provided better temporal modeling of the 12-lead ECG signal \cite{Gundlapalle2022SingleTo12Lead,Hebiguchi2025ChestLeadReconstruction}.
    \item Attention-based architectures: Attention-modified U-Net models have shown average Pearson correlations of around 0.80 and $R^2$ values of approximately 0.64 \cite{Garg2023AttentionUNetECG}.
    \item Lightweight architectures: Frequency-based decomposition and parameter-efficient 12 lead networks have been developed for use in resource-constrained environments (such as wearables) \cite{Lee2025mEcgNet}.
\end{itemize}

Recently, the ECG reconstruction have moved away from direct signal-to-signal mapping to training on channel-agnostic latent representation. These methods involve joint alignment and reconstruction across multiple leads or multimodal contrastive alignment such as integrating cardiac MRI and ECG data. These representations aim to preserve channel-specific information as well as facilitate information transfer across different lead subsets \cite{Ibtehaz2024ModallyReduced,Selivanov2025GlobalLocalContrastive}. Most existing studies are conducted on PTB Diagnostic Database \cite{Smith2021FTDNN,Gundlapalle2022SingleTo12Lead,Beco2022LeadConversion,Obianom2025ECGReconstruction,moghaddam2025,maheshwari2013}, which, despite its historical significance, is relatively small in terms of subject count and pathological diversity. In contrast, the more recent and extensive PTB-XL dataset remains underutilized \cite{Lu2025LightweightU2Net,11096411,chen2024mcma}, leaving a gap in evaluating how these models scale to larger, more heterogeneous clinical populations. Furthermore, performance is typically inflated by subject-dependent partitioning as intra-patient data is mixed across training and test sets \cite{Lu2025LightweightU2Net,Smith2021FTDNN,banta2021cnn_ecg,Garg2023AttentionUNetECG}. This leads to subject-specific memorization rather than clinical generalization. Only few studies have been done in patient-independent settings \cite{11096411,chen2024mcma}.

\section{Proposed Methodology}
The proposed methodology reformulates ECG lead reconstruction as a multi-view integration task, combining morphological fidelity (time-domain waveforms) with a pathology-aware latent space. The system comprises three stages: signal preprocessing, dual-representation construction, and stacked latent decoding.
\subsection{Signal Cleaning and Segmentation}
Each 12-lead ECG is processed through a narrowband notch filter, 0.5–45Hz Butterworth band-pass filter and Median-based filter eliminating powerline interference, baseline drift and residual low-frequency wander, respectively. Cleaned signals are segmented into temporal windows of $T=256$ samples with a 64-sample hop size. Segments are quality controlled based on lead-wise amplitude and root-mean-square criteria using empirical percentile bounds (0.1–99.9\%). Artifact-free segments $\mathbf{x} \in \mathbb{R}^{3\times256}$ (leads I, II, and $V_2$) serve as inputs for reconstructing leads $V_1$ and $V_3$–$V_6$.
\subsection{Pathology-aware Latent Representation}
To capture clinically meaningful invariants, we train a network $f_{\phi}:\mathbf{x}\rightarrow\mathbf{h}$ to structure a latent space where similar pathologies cluster together. Clean segments are transformed into augmented morphology-centered (R-peak aligned) and signal-augmented views, then processed by a 1-D convolutional encoder into an embedding $\mathbf{h} \in \mathbb{R}^{128}$ and projected into an $\ell_2$-normalized embedding $\mathbf{z} \in \mathbb{S}^{d-1}$.
We employ a supervised contrastive loss $\mathcal{L}_{\mathrm{sc}}$ with temperature $\tau = 0.07$ to maximize mutual information between the latent space and high-confidence diagnostic labels $l$ (certainty $\geq 80$):
\begin{equation}
\mathcal{L}_{\mathrm{sc}} = \frac{1}{|\mathcal{B}|} \sum_{i\in\mathcal{B}} \left[ -\frac{1}{|\mathcal{P}(i)|} \sum_{p\in\mathcal{P}(i)} \log \frac{ e^{\,\mathbf{z}_i^\top \mathbf{z}_p / \tau} }{ \sum_{j \in \mathcal{B} \setminus {i}} e^{\,\mathbf{z}_i^\top \mathbf{z}_j / \tau} } \right]
\label{eq:supcon}
\end{equation}
where $\mathcal{P}(i)$ contains indices of samples in batch $\mathcal{B}$ sharing at least one label with anchor $i$. $\mathcal{L}_{\mathrm{sc}}$ shapes the latent representation with high intra-class compactness and inter-class separation, providing a robust, pathology-aware anchor that can guide the downstream reconstruction decoder. All clean segments are projected into $\mathbf{h}$ after training $f_{\phi}$.
\subsection{Normalization and Fusion}
The time-domain waveform $\mathbf{x}$ and learned representation $\mathbf{h}$ are normalized to stabilize the reconstruction process. The segment $\mathbf{x}$ is normalized lead-wise to zero mean and unit variance: $\mathbf{\hat{x}}_{c,i} = (\mathbf{x}_{c,i} - \mathbf{\mu}_c) / (\sigma_c + \epsilon)$, where $\mu_c$ and $\sigma_c$ are calculated across the temporal dimension $T=256$. Similarly, $\mathbf{h}$ is normalized as $\mathbf{\hat{h}} = (\mathbf{h} - \mathbf{\mu}) / (\sigma + \epsilon)$ using the mean vector and standard deviation of $\mathbf{h}$.
\subsection{Stacked Latent Decoder}
\begin{figure}[htbp]
    \centering
    \includegraphics[width=\linewidth]{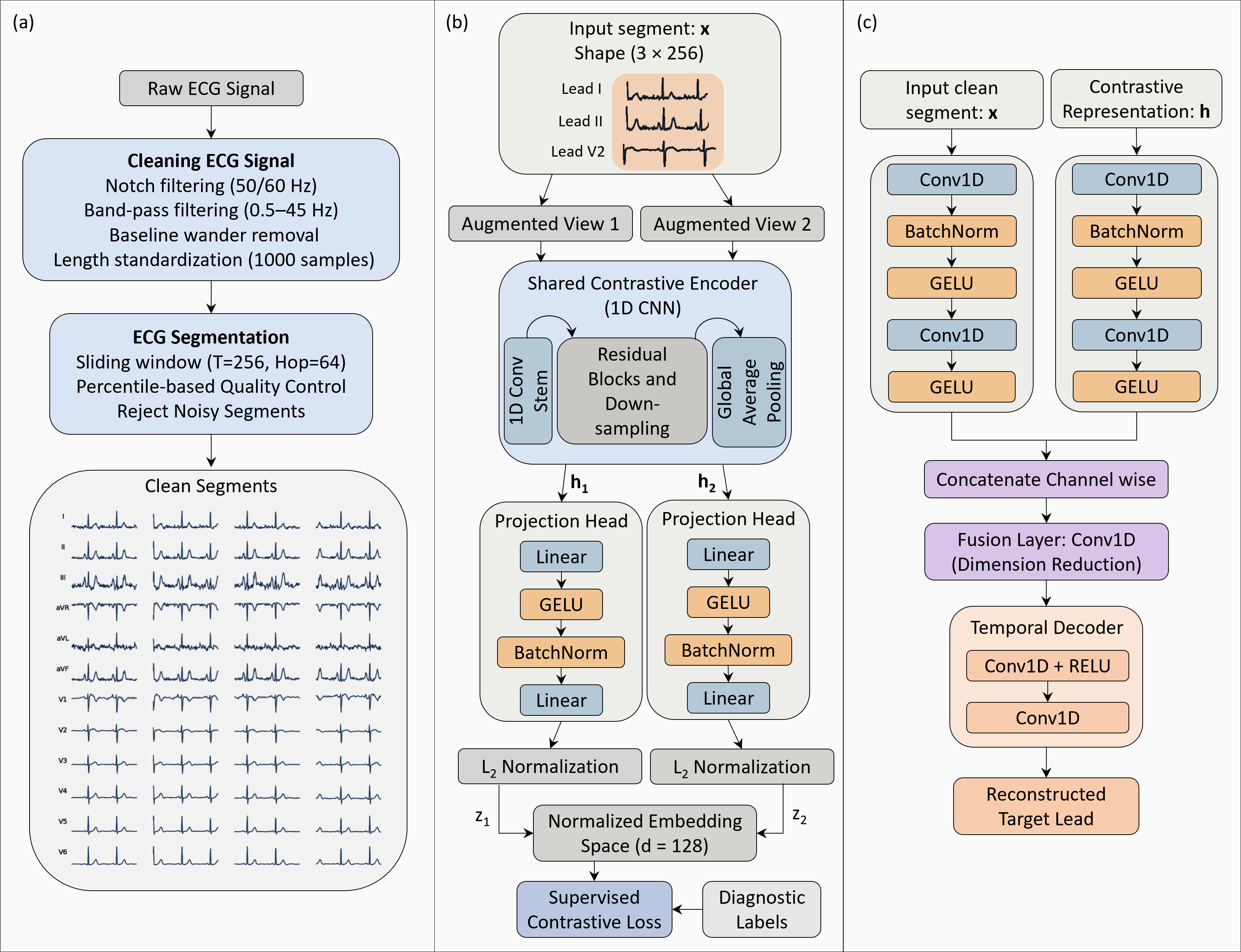}
    \caption{Proposed ECG reconstruction framework. (a) Preprocessing and segmentation, (b) Multi-View Supervised contrastive pretraining to learn pathology-aware latent representations, and (c) Reconstruction network that integrates contrastive representation with clean ECG signals via a temporal decoder.}
    \label{fig:final_model}
\end{figure}

The final reconstruction network $f_\theta:\mathbf{\hat{x}}, \mathbf{\hat{h}}\rightarrow \mathbf{y}$ synthesizes target lead by integrating $\mathbf{\hat{x}} \in \mathbb{R}^{3\times256}$ and $\mathbf{\hat{h}} \in \mathbb{R}^{128}$. Dedicated projection modules map both inputs to a shared 128-channel temporal dimensionality. The features are then concatenated into a stacked latent tensor $\in \mathbb{R}^{256\times256}$ and fused via a 1-D convolutional layer. A compact temporal decoder then recovers the target waveform $\mathbf{y} \in \mathbb{R}^{256}$. Independent decoders are trained for each output lead. The overall workflow of the methodology is illustrated in Fig.\ref{fig:final_model}. The Decoder loss function combines Mean Squared Error for sensitivity to large deviations and Mean Absolute Error to provide a steady gradient against outliers (CITE PAPER HERE).

\section{Experiments}
\subsection{Data Partitioning and Protocol}
We utilized PTB-XL dataset \cite{PhysioNet-ptb-xl-1.0.1} with a patient-wise partitioning strategy to prevent inter-subject leakage. First 8 folds were used for training, 9th for validation, and last for testing. For cross-dataset evaluation, we used PTB Dataset \cite{Goldberger2000PhysioNet}. All signals were resampled to 100 Hz for consistency.
\subsection{Training Configuration and Evaluation Metrics}
 Decoder is trained using AdamW and early stopping on validation loss. For testing, non-overlapping segments are utilized to reconstruct the complete 10-second signal, ensuring a fair and realistic estimation of RMSE in terms of clinical viability. Fidelity is quantified using Root Mean Squared Error (RMSE) in mV, the Coefficient of Determination ($R^2$), and the Pearson Correlation Coefficient \cite{moghaddam2025, Obianom2025ECGReconstruction, Sohn2020ReconstructionECG, Gundlapalle2022SingleTo12Lead, Garg2023AttentionUNetECG, Lu2025LightweightU2Net}.

\subsection{Latent Space Structure and Clinical Separability}
\begin{figure}[t]
\centering
\includegraphics[width=\columnwidth]{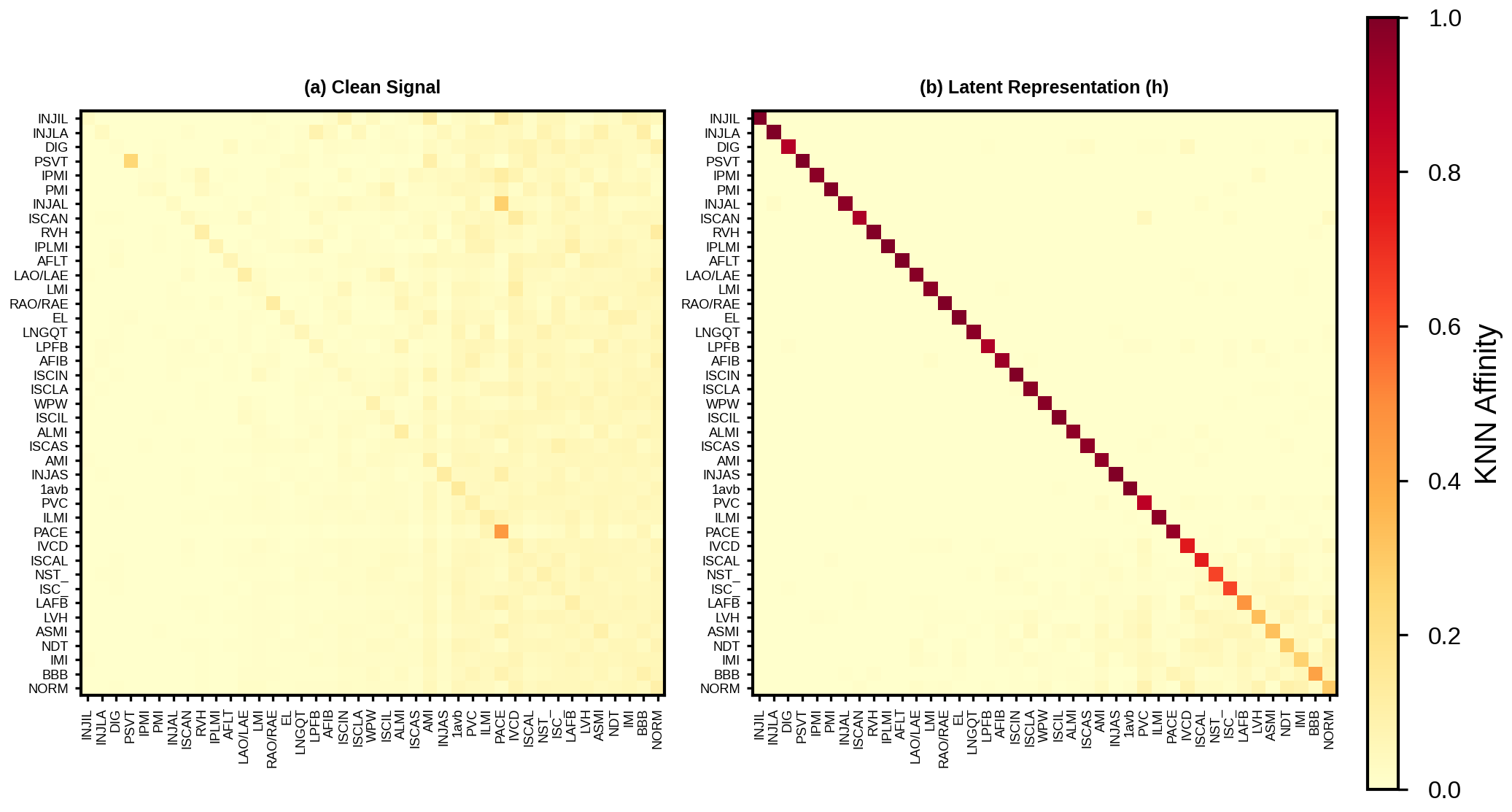}
\caption{k-NN affinity matrices ($k=10$) comparing (a) the clean baseline $\mathbf{x}$ and (b) the learned latent representation $\mathbf{h}$.}
\label{fig:pairwise_knn_analysis}
\end{figure}

To quantify clinical separability, we analyze the cosine similarity of test data identifying $k=10$ nearest neighbors of clean signal $\hat{\mathbf{x}}$ and latent embeddings $\hat{\mathbf{h}}$ and compute class-to-class affinity matrix A, where element $A_{i,j}$ is the mean proportion of neighbors for class $i$ that belongs to class $j$. Fig. \ref{fig:pairwise_knn_analysis} clearly shows that the affinity for latent $\mathbf{h}$ is stronger along the diagonal than the original clean baseline signal $\mathbf{x}$. The diagonal average (intra-class consistency) for $\mathbf{h}$ increased to 0.837 compared to 0.083 in $\mathbf{x}$, demonstrating that the supervised contrastive loss effectively partitions the latent space in pathology-aware clusters.
\subsection{Lead Reconstruction Performance}
\begin{figure}[t]
\centering
\includegraphics[width=\columnwidth]{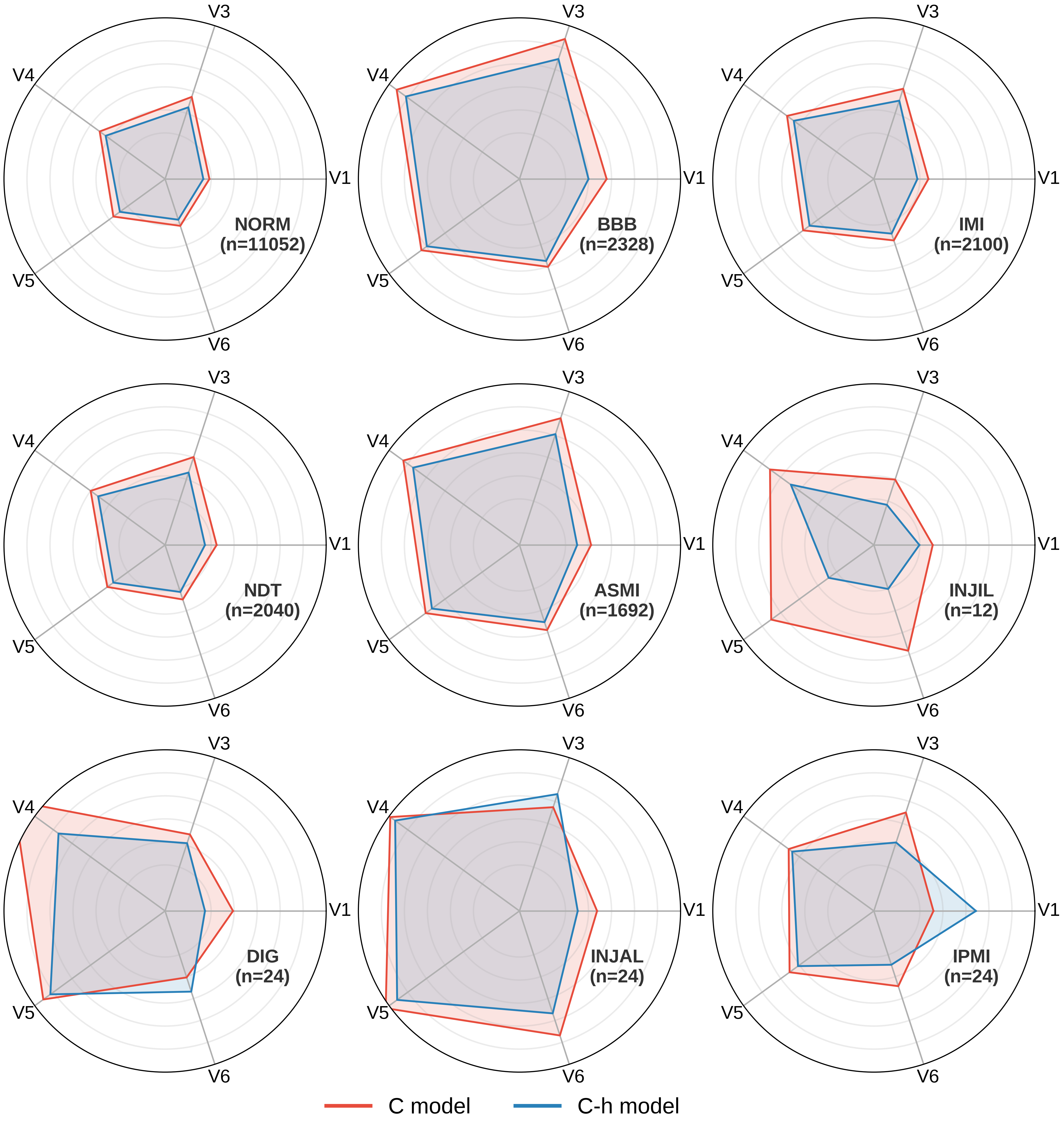}
\caption{Radar plot showing lower RMSE for C-h model for few most and least abundant diagnostic class. $n$ represents the number of segments in test fold. The closer the vertex to center, the lower is the RMSE. The y-values for each ring correspond to [0.05, 0.075, 0.10, 0.125, 0.15] starting from center.}
\label{fig:rmse_heatmap}
\end{figure}

\begin{figure}[t]
    \centering
    \begin{subfigure}[t]{0.32\linewidth}
        \centering
        \includegraphics[width=\linewidth]{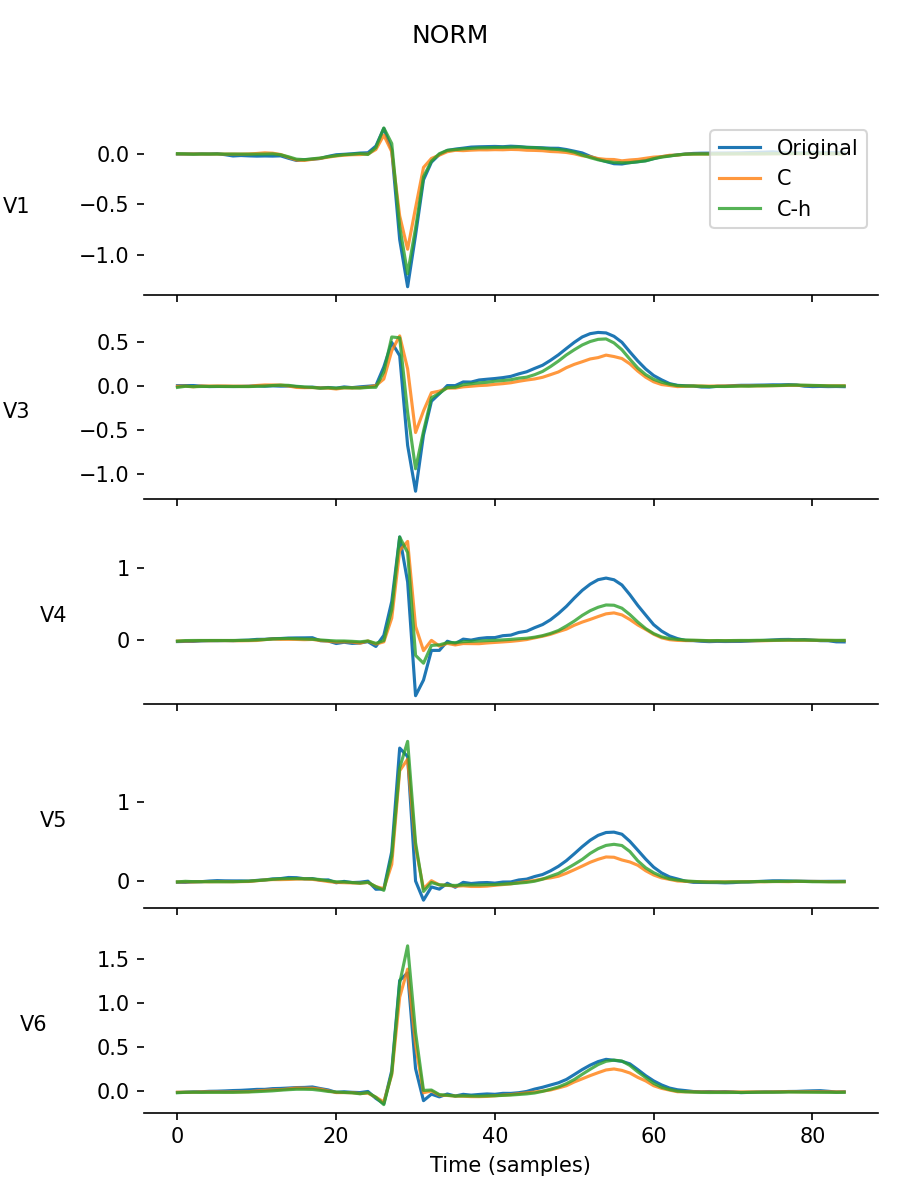}
    \end{subfigure}
    \hfill
    \begin{subfigure}[t]{0.32\linewidth}
        \centering
        \includegraphics[width=\linewidth]{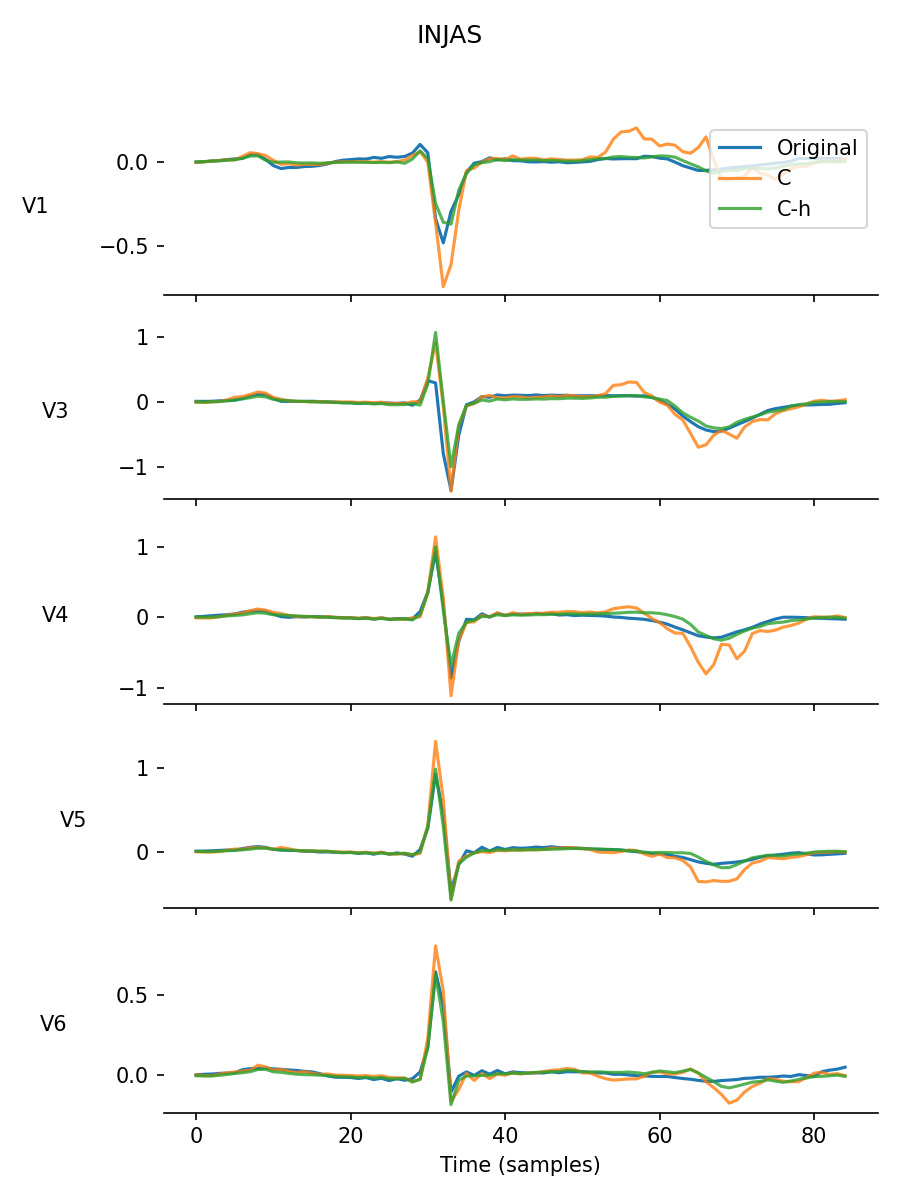}
    \end{subfigure}
    \hfill
    \begin{subfigure}[t]{0.32\linewidth}
        \centering
        \includegraphics[width=\linewidth]{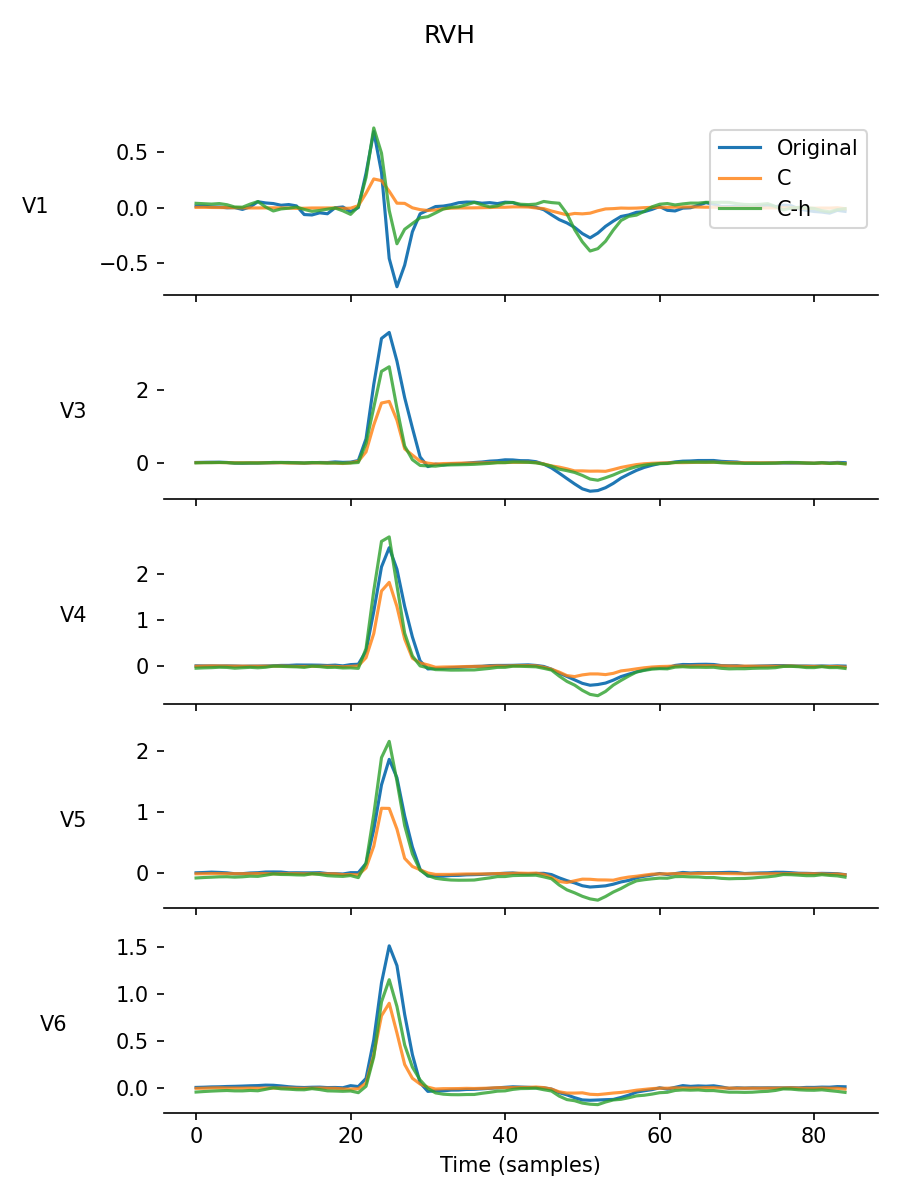}
    \end{subfigure}

    \caption{Qualitative heartbeat-level reconstruction examples across different diagnostic classes (NORM, INJAS, RVH) for leads V1,V3--V6. The C-h configuration more closely follows the ground truth compared to the C baseline, particularly in capturing QRS complex amplitude and T-wave morphology.}
    \label{fig:reconstruction_examples}
\end{figure}

The impact of the pathology-aware anchor on signal fidelity is evaluated across all target leads for the test fold. As detailed in Table~\ref{tab:reconstruction_results}, the proposed Clean + $\mathbf{h}$ model (C-h) outperforms the baseline model (C) with Clean-only signal across all metrics. Notable improvements are observed in the transitional leads ($V_3$–$V_4$), with $R^2$ increasing by approximately 11.0\% and 13.5\%, respectively. This suggests that the latent anchor effectively restricts the solution space to a pathological manifold, allowing the decoder to recover morphological features typically lost to population-wide averaging. Quantitative robustness across pathologies is further analyzed via radar plot for four most and least abundant diagnostic classes (Fig.~\ref{fig:rmse_heatmap}). The C-h model maintains consistently lower error rates compared to the baseline, even for rare pathologies, indicating that supervised contrastive alignment anchors the reconstruction to disease-specific invariants rather than merely optimizing for the majority class. Qualitative results (Fig.~\ref{fig:reconstruction_examples}) also confirm that C-h model achieves superior alignment with ground truth waveforms compared to C model.

\begin{table}[t]
\caption{Reconstruction Performance: Baseline (C) vs. Proposed (C-h)}
\label{tab:reconstruction_results}
\centering
\scriptsize
\setlength{\tabcolsep}{3pt}
\begin{tabular}{l ccc ccc ccc}
\hline
\textbf{Lead} & \multicolumn{3}{c}{\textbf{RMSE (mV) $\downarrow$}} & \multicolumn{3}{c}{\textbf{$R^2$ $\uparrow$}} & \multicolumn{3}{c}{\textbf{Pearson $\uparrow$}} \\
\cline{2-4} \cline{5-7} \cline{8-10}
& \textbf{C} & \textbf{C-h} & \textbf{$\Delta\%$} & \textbf{C} & \textbf{C-h} & \textbf{$\Delta\%$} & \textbf{C} & \textbf{C-h} & \textbf{$\Delta\%$} \\
\hline
$V_1$ & 0.063 & \textbf{0.052} & \textcolor{darkgreen}{+18.2\%} & 0.754 & \textbf{0.839} & \textcolor{darkgreen}{+11.2\%} & 0.940 & \textbf{0.951} & \textcolor{darkgreen}{+1.2\%} \\
$V_3$ & 0.115 & \textbf{0.101} & \textcolor{darkgreen}{+12.7\%} & 0.670 & \textbf{0.744} & \textcolor{darkgreen}{+11.0\%} & 0.913 & \textbf{0.926} & \textcolor{darkgreen}{+1.5\%} \\
$V_4$ & 0.118 & \textbf{0.108} & \textcolor{darkgreen}{+8.7\%} & 0.605 & \textbf{0.687} & \textcolor{darkgreen}{+13.5\%} & 0.893 & \textbf{0.906} & \textcolor{darkgreen}{+1.4\%} \\
$V_5$ & 0.094 & \textbf{0.084} & \textcolor{darkgreen}{+10.3\%} & 0.684 & \textbf{0.717} & \textcolor{darkgreen}{+4.8\%} & 0.922 & \textbf{0.931} & \textcolor{darkgreen}{+1.0\%} \\
$V_6$ & 0.071 & \textbf{0.063} & \textcolor{darkgreen}{+11.4\%} & 0.699 & \textbf{0.760} & \textcolor{darkgreen}{+8.8\%} & 0.935 & \textbf{0.944} & \textcolor{darkgreen}{+1.0\%} \\
\hline
\textbf{Mean} & 0.092 & \textbf{0.082} & \textcolor{darkgreen}{\textbf{+12.2\%}} & 0.683 & \textbf{0.749} & \textcolor{darkgreen}{\textbf{+9.9\%}} & 0.921 & \textbf{0.932} & \textcolor{darkgreen}{\textbf{+1.2\%}} \\
\hline
\end{tabular}
\end{table}

\begin{table}[t]
\centering
\caption{Record-level Comparison with Rajotte \textit{et al.} \cite{11096411}}
\label{tab:final_comparison}
\scriptsize
\setlength{\tabcolsep}{2pt}
\resizebox{\columnwidth}{!}{
\begin{tabular}{l|cc|cccc}
\hline
\textbf{Lead} & \multicolumn{2}{c|}{\textbf{Proposed (C-h)}} & \multicolumn{4}{c}{\textbf{Rajotte \textit{et al.} \cite{11096411}}} \\
\cline{2-7}
 & RMSE & $R^2$ & RMSE & \textbf{\color{darkgreen}$\Delta RMSE \%$} & $R^2$ & \textbf{\color{darkgreen}$\Delta R^2 \%$} \\
\hline
$V_1$ & $0.054 \pm 0.002$ & $0.839 \pm 0.010$ & $0.276 \pm 0.009$ & \color{darkgreen}80.4\% & $0.796 \pm 0.011$ & \color{darkgreen}5.4\% \\
$V_3$ & $0.103 \pm 0.003$ & $0.747 \pm 0.023$ & $0.364 \pm 0.011$ & \color{darkgreen}71.7\% & $0.702 \pm 0.023$ & \color{darkgreen}6.4\% \\
$V_4$ & $0.110 \pm 0.004$ & $0.692 \pm 0.026$ & $0.420 \pm 0.014$ & \color{darkgreen}73.8\% & $0.613 \pm 0.040$ & \color{darkgreen}12.9\% \\
$V_5$ & $0.086 \pm 0.003$ & $0.719 \pm 0.037$ & $0.350 \pm 0.011$ & \color{darkgreen}75.4\% & $0.694 \pm 0.023$ & \color{darkgreen}3.6\% \\
$V_6$ & $0.065 \pm 0.002$ & $0.765 \pm 0.028$ & $0.334 \pm 0.012$ & \color{darkgreen}80.5\% & $0.729 \pm 0.018$ & \color{darkgreen}4.9\% \\
\hline
Mean & \textbf{0.084} & \textbf{0.752} & 0.349 & \textbf{\color{darkgreen}76.4\%} & 0.707 & \textbf{\color{darkgreen}6.6\%} \\
\hline
\end{tabular}
}
\end{table}

\subsection{Comparison with Prior Work}
Very few studies strictly adhere to a patient-independent evaluation protocol. Many existing frameworks, such as those by Lu \textit{et al.} \cite{Lu2025LightweightU2Net} and Garg \textit{et al.} \cite{Garg2023AttentionUNetECG}, utilize random record-level splits where ECG from same subject may appear in both training and test sets, thereby inflating results. In contrast, we adopt a rigorous patient-wise partition to ensure realistic clinical generalization. We primarily contextualize our results against Rajotte \textit{et al.} \cite{11096411}, which utilized patient-wise partition on PTB-XL dataset with input leads I, II, and $V_2$. Also, Rajotte \textit{et al.} pretrained a CNN+BiLSTM model on PTB dataset first before fine-tuning on PTB-XL and reconstructed full 10s signal. To ensure a fair comparison we adopted a record-level evaluation protocol by reassembling our 2.56-second reconstructed segments into full 10-second signals. The proposed C-h model achieves statistically robust improvements across all precordial leads (Table~\ref{tab:final_comparison}), achieving an average RMSE reduction of 76.4\% and 6.6\% improvement in $R^2$ compared to Rajotte \textit{et al.} The proposed model capacity also remains modest, with $\sim239K$ parameters, which is comparable to prior work ($\sim252K$ parameters)~\cite{11096411}.

Further, our method shows superior Pearson correlation (0.932) compared to existing models that often utilize random splits, as contextualized in Table~\ref{tab:comparison_recon_five_papers}. These results underscore that anchoring reconstruction to a pathology-aware latent space significantly outperforms unconditioned, direct-mapping models.

\begin{table}[t]
\centering
\caption{Global Performance Benchmarks on PTB-XL}
\label{tab:comparison_recon_five_papers}
\scriptsize
\setlength{\tabcolsep}{3pt}
\begin{tabular}{lcccc}
\hline
Method & Patient Indep. & Input Leads & RMSE & Pearson \\
\hline
Garg \textit{et al.} (2023) \cite{Garg2023AttentionUNetECG} & No & II & – & 0.805 \\
Lu \textit{et al.} (2025) \cite{Lu2025LightweightU2Net} & No & $V_2$ & 0.160 & 0.860 \\
Hebiguchi \textit{et al.} (2025) \cite{Hebiguchi2025ChestLeadReconstruction} & No & I, II, $V_2$ & - & 0.93 \\
\hline
\textbf{Ours} & \textbf{Yes} & I, II, $V_2$ & \textbf{0.082} & \textbf{0.932} \\
\hline
\end{tabular}
\end{table}

\begin{table}[H]
\centering
\caption{Cross-Dataset Performance Comparison on the PTB Database}
\label{tab:lead_reconstruction_comparison_ptb}
\scriptsize
\setlength{\tabcolsep}{3pt}
\begin{tabular}{lcccc}
\hline
Method & Patient Indep. & Input Leads & RMSE & Pearson \\
\hline
Smith \textit{et al.} (2021) \cite{Smith2021FTDNN} & No & 6 limb leads + $V_2$ & 0.173 & \textbf{0.921} \\
Moghaddam \textit{et al.} (2025) \cite{moghaddam2025} & Yes & 3-leads & – & 0.76 \\
\hline
\textbf{Ours} & \textbf{Yes} & I, II, $V_2$ & \textbf{0.095} & 0.887 \\
\hline
\end{tabular}
\end{table}


\subsection{Cross-Dataset Generalization}
To evaluate the robustness of the pathology-aware representation $\mathbf{h}$ under distribution shifts, we perform a cross-dataset evaluation on the independent PTB Diagnostic ECG Database. Unlike prior benchmarks that are typically trained and tested on the same dataset, our model is trained exclusively on PTB-XL and evaluated on PTB without subject-specific calibration or fine-tuning. Table~\ref{tab:lead_reconstruction_comparison_ptb} summarizes the comparison of our method with other patient-independent metrics reported in the literature on PTB dataset. Despite the domain shift and reduced input set, our proposed C-h model achieves an average RMSE of $0.095$~mV, representing a substantial improvement over the $0.173$~mV reported by Smith \textit{et al.}. Furthermore, our framework attains an average Pearson correlation of $0.887$ without training on PTB dataset. These results empirically validate that anchoring the reconstruction to a disease-aware latent space allows synthesized waveforms to maintain clinically consistent morphologies that generalize across different recording environments and hardware.

\section{Conclusion}

In conclusion, the proposed framework achieves a strong balance between accuracy and efficiency. Despite its compact size, it attains low reconstruction error with average RMSE of $0.082$ on PTB-XL under a patient-independent setting and maintains good generalization across datasets with average RMSE of $0.095$ on the PTB database. The model operates with very low inference latency of approximately $0.035$ ms, enabling real-time deployment.

By leveraging pathology-aware representations, the proposed approach preserves clinically relevant ECG morphology while remaining computationally efficient, making it suitable for practical and portable diagnostic applications.

\bibliographystyle{unsrt}
\bibliography{references}
\end{document}